\documentclass[runningheads]{llncs}
\usepackage{mathtools}
\usepackage{hyperref}
\usepackage{url}
\usepackage{graphicx}
\usepackage{subcaption}
\usepackage{xcolor}
\usepackage{tabularx}
\usepackage{comment}
\usepackage[export]{adjustbox}
\usepackage[ruled, linesnumbered]{algorithm2e}
\begin{document}
\title{A Comprehensive Approach to Unsupervised Embedding Learning based on AND Algorithm
}
%
%\titlerunning{Abbreviated paper title}
% If the paper title is too long for the running head, you can set
% an abbreviated paper title here
%
\author{Sungwon Han\inst{1} \and
Yizhan Xu\inst{2} \and
Sungwon Park\inst{1}  \and \\
Meeyoung Cha\inst{3,1}  \and
Cheng-Te Li\inst{2}
}

\authorrunning{S. Han et al.}

\institute{School of Computing, KAIST \and Institute of Data Science, National Cheng Kung University \and Data Science Group, Institute for Basic Science}

\maketitle              % typeset the header of the contribution
\begin{abstract}
Unsupervised embedding learning aims to extract good representation from data without the need for any manual labels, which has been a critical challenge in many supervised learning tasks. This paper proposes a new unsupervised embedding approach, called Super-AND, which extends the current state-of-the-art model~\cite{huang2019unsupervised}. Super-AND has its unique set of losses that can gather similar samples nearby within a low-density space while keeping invariant features intact against data augmentation. Super-AND outperforms all existing approaches and achieves an accuracy of 89.2\% on the image classification task for CIFAR-10. We discuss the practical implications of this method in assisting semi-supervised tasks. 
\end{abstract}

\section{Introduction}
Unsupervised embedding learning aims to extract visually meaningful representations without any label information. Here ``visually meaningful'' refers to finding features that satisfy two traits: (\textit{1}) positive attention and (\textit{2}) negative separation~\cite{oh2016deep,ye2019unsupervised}. Data samples from the same class (i.e., positive samples) should be close to one another in the embedding space. In contrast, those from different classes (i.e., negative samples) should be positioned far away from each other. This task is challenging in unsupervised learning since a model does not know a priori which data points are positive or negative.

Several innovative approaches have been proposed. The sample specificity approach starts by assuming all data points to be negative samples and separates them in the feature space~\cite{wu2018unsupervised,bojanowski2017unsupervised}. This approach, however, suffers from a bias that the learning process relies solely on negative separation. The data augmentation approach, in contrast, considers positive samples in training~\cite{ye2019unsupervised}. This new approach efficiently reduces the ambiguity in supervision while keeping invariant features in the embedding space. More recently, the anchor neighborhood discovery (AND) method has been proposed, which alleviates the complexity in boundaries by progressively discovering the nearest neighbor among the data points~\cite{huang2019unsupervised}. These methods each tackle the different limitations of the sample specificity approach. However, no unified approach exists.

This paper presents a holistic method for unsupervised embedding learning, named Super-AND. It extends the AND~\cite{huang2019unsupervised} algorithm and unifies the merits of previous methods with its unique architecture. Our model not only focuses on learning distinctive features across neighborhoods but also utilizes the edge information in embeddings to maintain the invariant class properties in data augmentation. The model also newly introduces the Unification Entropy loss (UE-loss), an adversary of the sample specificity loss that is critical in gathering similar data points within a low-density space. Extensive experiments on several benchmark datasets verify the superiority of Super-AND. The main contributions of this paper are as follows:

%The results show the synergetic advantages among modules of Super-AND. 

\begin{itemize}
    \item We propose a comprehensive model, Super-AND, which unifies some of the key techniques from existing research and introduces a novel loss (UE-loss) that gathers similar samples in a low-density space.
    \smallskip
    \item For the image classification task, Super-AND outperforms all baselines with an accuracy of 89.2\% in CIFAR-10 with the ResNet18 backbone network,  which is a 2.9\% gain over the state-of-the-art. 
    \smallskip
    \item The ablation study shows that every component of Super-AND contributes to the performance increase and that their synergy is critical.
    \smallskip
    \item Super-AND can further improve better-understood tasks like semi-supervised learning when used in the pre-training step.
\end{itemize}

The outstanding performance of Super-AND is a step closer to the broader adoption of unsupervised techniques in computer vision tasks. The premise of ``data-less embedding learning'' is at its applicability to practical scenarios, where there exists only one or two examples per cluster. Codes and training data from this research are accessible via the github link\footnote{https://github.com/super-AND/super-AND}.

\if 0
\begin{figure}[ht!]
\centering
  \begin{subfigure}[b]{0.28\textwidth}
    \includegraphics[width=\textwidth]{figure/Positive_concentration.png}
    \caption{Positive attention}
    \label{fig:intro_1} 
  \end{subfigure}
  \hspace{10mm}
  \begin{subfigure}[b]{0.28\textwidth}
    \includegraphics[width=\textwidth]{figure/Negative_separation.png}
    \caption{Negative separation}
    \label{fig:intro_2}
  \end{subfigure}
  \caption{Illustration of two characteristics: (a) Positive attention, (b) Negative separation. Data points with same shape and same color in this plot are positive samples. Otherwise, they are negative samples.}
\end{figure}
\fi
\section{Related Work}
Existing studies can be summarized into four groups. 
\smallskip

\noindent
\textbf{Generative model.} This type of model is a powerful branch in unsupervised learning. By reconstructing data based on its underlying distribution, a model can generate new data points as well as features without label supervision. Generative adversarial networks (GAN)~\cite{goodfellow2014generative} have led rapid progress in image generation problems~\cite{arjovsky2017wasserstein,zhang2019self}. Several studies have utilized GAN for unsupervised embedding learning~\cite{radford2015unsupervised}. However, the main objective of generative models lies at mimicking the true distribution of each class, rather than discovering distinctive categorical information the data contains. 
\smallskip

\noindent
\textbf{Self-supervised learning.} This type of learning uses inherent structures in images as pseudo-labels and exploits labels for back-propagation. For example, a model can be trained to create embeddings by predicting the relative position of a pixel from other pixels~\cite{doersch2015unsupervised} or the degree of changes after rotating images~\cite{gidaris2018unsupervised}. Predicting future frames of a video can benefit from this technique~\cite{walker2016uncertain}. All of these methods are suitable for unsupervised embedding learning, although there exists a risk of false knowledge from generated labels that weakly correlate with the underlying class information.
\smallskip

\noindent
\textbf{Sample-specificity learning.} This type of model learns feature representation from capturing apparent discriminability among instances. By considering all instances as a single individual class, they trained the classifier to separate data points in a confined space~\cite{bojanowski2017unsupervised,wu2018unsupervised}. This learning method holds because a deep learning model can detect the similarity through supervised learning. However, its decision boundary becomes complex when the number of classes gets larger.
\smallskip

\noindent
\textbf{Clustering analysis.} This type of analysis is an extensively studied area in unsupervised learning, whose main objective is to group similar objects into the same class. Many studies either leverage deep learning for dimensionality reduction before clustering~\cite{schroff2015facenet,baldi2012autoencoders} or train models in an end-to-end fashion~\cite{xie2016unsupervised,yang2016joint}. Deep cluster~\cite{caron2018deep} is an iterative method that updates its weights by predicting cluster assignments as pseudo-labels. However, directly reasoning the global structures without any label is known to be error-prone. The AND model~\cite{huang2019unsupervised}, which we extend in this study, combines the advantages of the sample specificity approach and the clustering method to mitigate any noisy supervision via neighborhood analysis.

%\noindent
%\textbf{Learning invariants from augmentation.} Data augmentation is a strategy that enables a model to learn from datasets with an increased variety of instances. These techniques do not deform any crucial features of data, but only change the style of images. Some studies hence, use augmentation techniques and train models to learn invariant features. In particular,~\cite{iic_arxiv} used mutual information to extract invariant features between the augmented images, and ~\cite{ye2019unsupervised} regarded the augmented images as positive samples of original pictures for unsupervised feature learning. This study also adopts the same concept to reduce the distance of relationship vectors between the original and the augmented images.
%\smallskip

\section{The Super-AND Model}

\begin{figure}[t!]
\centering
%\hspace{-12mm}
\includegraphics[width=1\textwidth]{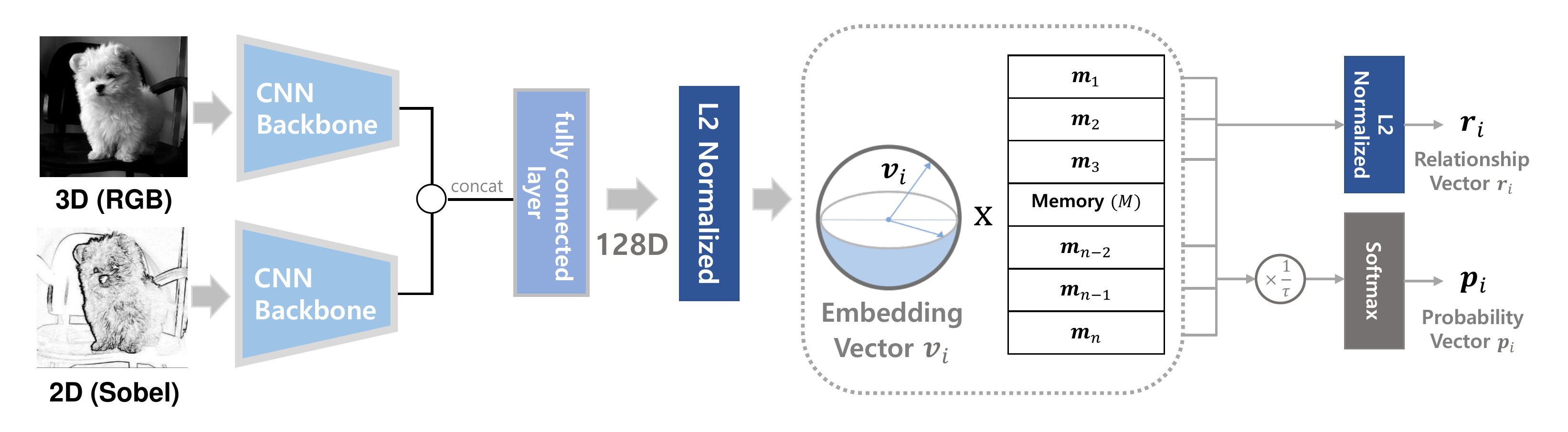}
\caption{Illustration of basic architecture in Super-AND.} 
%\vspace{-3mm}
\label{fig:basic_model} 
\end{figure}
\textbf{Problem definition.} Assume that we are given an unlabeled image set $\mathcal{I}$, and a batch set $\mathcal{B}$ with $n$ images: $\mathcal{B} = \{\mathbf{x}_1, \mathbf{x}_2, ..., \mathbf{x}_n\} \subset \mathcal{I}$. The goal of unsupervised embedding learning is to obtain a feature extractor $f_\theta$ whose generated representations (i.e., $\mathbf{v}_i = f_\theta (\mathbf{x}_i)$) are ``visually meaningful,'' a definition we discussed earlier.

Let $\mathcal{\hat{B}} = \{\mathbf{\hat{x}}_1, \mathbf{\hat{x}}_2, ..., \mathbf{\hat{x}}_n\}$ be the augmentation set of input batches $\mathcal{B}$. Super-AND projects images $\mathbf{x}_i$, $\mathbf{\hat{x}}_i$ from batches $\mathcal{B}$, $\mathcal{\hat{B}}$ to 128-dimensional embeddings $\mathbf{v}_i$, $\mathbf{\hat{v}}_i$. During this process, the Sobel-processed images~\cite{maini2008study} are used, and the feature vectors from both images are concatenated to emphasize the edge information in embeddings (i.e., see the left side in Fig.~\ref{fig:basic_model}). Concatenated vectors are projected to 128-dimensional sphere embeddings with L2 normalization, and these are treated as latent representations of given images.

Similar to the sample specificity learning~\cite{bojanowski2017unsupervised,wu2018unsupervised}, our model considers every data sample as an independent class at first. The model computes the probability $\mathbf{p}_i^j$ of the $i$-th image being recognized as the $j$-th class. Since every instance is a prototype of its class, the similarity between instance $i$ and class $j$ becomes a dot product of feature vectors ($\mathbf{v}_i$, $\mathbf{v}_j$) from two images $\mathbf{x}_i$ and $\mathbf{x}_j$. By applying the softmax function, we can calculate $\mathbf{p}_i$. We denote the superscript $j$ in vector notation as the $j$-th component value of a given vector. (Eq.~\ref{eq:non_param_before}).
\begin{align}
P(j|\mathbf{v}_i) = \mathbf{p}_i^j &= \frac{{\exp (\mathbf{v}_j^{\top} \mathbf{v}_i)}}{\sum_{k=1}^n \exp (\mathbf{v}_k^{\top} \mathbf{v}_i)} \label{eq:non_param_before}
\end{align}

Calculating embedding from all images (i.e., $\mathbf{v}_k$) in every iteration loop is computationally demanding. To reduce such complexity, we set up a memory bank $M$ to save an instance embedding $\mathbf{m}_i$ accumulated from the previous steps, as proposed in~\cite{wu2018unsupervised}. For every iteration, the $i$-th instance's embedding vector $\mathbf{m}_i$ in memory bank $M$ is updated by the exponential moving average~\cite{lucas1990exponentially}, taking into account the latest embedding vector $\mathbf{v}_i$ (Eq.~\ref{eq:memory_update}). With the help of embedding $m_i$ in the memory, the lightweight version of $\mathbf{p}_i$ can be calculated as in Eq.~\ref{eq:non_param}. The temperature parameter ($\tau < 1$) is added to ensure a label distribution with low entropy~\cite{hinton2015distilling}.  
\begin{align}
\mathbf{m}_i &= (1 - \eta) \cdot \mathbf{m}_i + \eta \cdot \mathbf{v}_i \label{eq:memory_update} \\ 
P(j|\mathbf{v}_i) &= \mathbf{p}_i^j = \frac{{\exp (\mathbf{m}_j^{\top} \mathbf{v}_i / \tau)}}{\sum_{k=1}^n \exp (\mathbf{m}_k^{\top} \mathbf{v}_i / \tau)} \label{eq:non_param}
\end{align}
\vspace*{-5mm}
\begin{align}
\mathbf{r}_i^j = \frac{\mathbf{v}_i \cdot \mathbf{m}_j}{||\mathbf{v}_i|| \cdot ||\mathbf{m}_j||},
  \ \ \  \mathbf{\hat{r}}_i^j = \frac{\mathbf{\hat{v}}_i \cdot \mathbf{m}_j}{||\mathbf{\hat{v}}_i|| \cdot ||\mathbf{m}_j||}
\label{eq:representation}
\end{align}

To detect any discrepancy after data augmentation, we define the relationship vectors $\mathbf{r}_i$, $\hat{\mathbf{r}_i}$ through the cosine similarity between the embedding vectors of original images and their augmented versions ($\mathbf{v}_i$, $\mathbf{\hat{v}}_i$), and the memory bank $M$ (Eq.~\ref{eq:representation}). Using $\mathbf{p}$, $\mathbf{v}$, and $\mathbf{r}$, we construct three losses: the AND-loss ($L_{and}$), the UE-loss ($L_{ue}$), and the AUG-loss ($L_{aug}$). The proposed model is finally trained by optimizing $L_{total}$ with the hyper-parameter $w(t)$ that controls the weights of UE-loss (Eq.~\ref{eq:w}). Below we describe each loss function.
\begin{align}
& L_{total} = L_{and} + w(t) \times L_{ue} + L_{aug} \label{eq:w}
\end{align}

\subsection{The AND-loss: Neighborhood discovery and supervision loss}
Existing clustering methods~\cite{caron2018deep,xie2016unsupervised} train networks to find an optimal mapping. However, their learning process is unstable due to random initialization and overfitting~\cite{zhang2017deep}. The newly proposed Anchor Neighbourhood Discovery (AND) method suggests a finer-grained clustering that gradually discovers `neighborhoods'~\cite{huang2019unsupervised}. By considering the nearest-neighbor pairs as local classes, AND can separate data points that belong to different neighborhood sets from those in the same neighborhood set. We adopt this novel discovery strategy for Super-AND.

The AND algorithm has three main steps: (1) neighborhood discovery, (2) progressive neighborhood selection with curriculum, and (3) neighborhood supervision. For the first step, the $k$ nearest neighborhood ($k$-NN) algorithm is used to discover all neighborhoods (Eq.~\ref{eq:3} and Eq.~\ref{eq:4}). When $k$ is set to 1, each neighborhood constructs the pair of two nearest data points.
\begin{align}
&\Tilde{\mathcal{N}} (\mathbf{x}) = \{\mathbf{x}_i | \mathbf{x}_i \neq \mathbf{x},\ f_{\theta}(\mathbf{x}_i)^{\top} f_{\theta}(\mathbf{x})\text{ is top-$k$ in } \mathcal{I}\} \label{eq:3}  \\
&\Tilde{\mathcal{N}} = \{(\mathbf{x}_i, \Tilde{\mathcal{N}} (\mathbf{x}_i))\ |\  \mathbf{x}_i \in \mathcal{I}\}  \label{eq:4} 
\end{align}

The neighbor pairs are then progressively selected for curriculum learning over multiple rounds. Every round, the model gradually increases the number of discovered neighborhoods for training. The key here is to keep the discovery of multiple class ``consistent'' clusters, where its progressive nature helps provide a consistent view of local class information at each round during training.

Selecting candidate neighborhoods for local classes rely on the entropy of probability vector $H(\mathbf{x}_i)$ (Eq.~\ref{eq:5}). Probability vector $\mathbf{p}_i$, obtained from the softmax function (Eq.~\ref{eq:non_param}), represents the visual similarity between training instances in a probabilistic manner. Data points with low entropy represent that they reside in a relatively low-density area and have only a few surrounding neighbors. Neighborhood pairs containing such data points tend to share consistent and easily distinguishable features, compared to other pairs. AND selects neighborhood set $\mathcal{N}$ from $\Tilde{\mathcal{N}}$ that is in a lower entropy order. Data points that are not selected belong to the complement set $\mathcal{N}^c$.
\begin{align}
&H(\mathbf{x}_i) = -\sum_j \mathbf{p}_i^j \log \mathbf{p}_i^j  \label{eq:5}
\end{align}

The AND-loss distinguishes neighborhood pairs from one another. Data points from the same neighborhoods need to be classified in the same class (i.e., the left-hand term in Eq.~\ref{eq:6}). As for the remaining data points that are not chosen, each forms an independent class (i.e., the right-hand term in Eq.~\ref{eq:6}).
\begin{align}
& L_{and} = - \sum_{i \in (\mathcal{B} \cap \mathcal{N})} \log (\sum_{j \in \{\mathbf{x}_i,\ \Tilde{\mathcal{N}} (\mathbf{x}_i)\}} \mathbf{p}_i^j) - \sum_{i \in (\mathcal{B} \cap \mathcal{N}^{\mathsf{c}})} \log \mathbf{p}_i^i \label{eq:6}
\end{align}

\subsection{The UE-loss: Unification entropy loss}
Existing sample specificity methods~\cite{wu2018unsupervised,bojanowski2017unsupervised} consider every data point as a prototype for a class. They use the cross-entropy loss to separate all data points in the L2-normalized embedding space. Due to the confined space, data points cannot be placed far away from one another, and similar samples are located nearby.

The unification entropy loss aims to strengthen the concentration-effect above. We define it as the entropy of the probability vector $\mathbf{\Tilde{p}}_i$. Probability vector $\mathbf{\Tilde{p}}_i$ is calculated from the softmax function, and this represents the similarity between instances except for every instance itself (Eq.~\ref{eq:ue_entropy}). By excluding the self-class, minimizing this loss leads nearby data points to approach each other even more aggressively. This idea is contrary to minimizing the sample specificity loss. Employing both the AND-loss and the UE-loss will enforce similar neighborhoods to be positioned close while keeping the overall neighborhoods as separated as possible. This loss is calculated as in Eq.~\ref{eq:ue_loss}.
\begin{align}
&\mathbf{\Tilde{p}}_i^j = \frac{{\exp (\mathbf{m}_j^{\top} \mathbf{v}_i / \tau)}}{\sum_{k=1, k \neq i}^n \exp (\mathbf{m}_k^{\top} \mathbf{v}_i / \tau)} \ \ \ \ \ \ \ \ \ \Tilde{H}(\mathbf{x}_i) = -\sum_{j \neq i} \mathbf{\Tilde{p}}_i^j \log \mathbf{\Tilde{p}}_i^j  \label{eq:ue_entropy} 
\end{align}
\vspace{-5mm}
\begin{align}
    L_{ue} = \sum_{i} \Tilde{H}(\mathbf{x}_i)  \label{eq:ue_loss}
\end{align}

\subsection{The AUG-loss: Data augmentation loss}
Data augmentation is one of the unsupervised methods to learn which features are substantial~\cite{iic_arxiv,ye2019unsupervised}. The premise of this method lies in the assumption that simple augmentation does not deform the underlying data characteristic. Then, invariant features learned from the augmented data should still contain class-relevant information, and hence a training network will show performance gain. 

The augmentation loss represents invariant image features. Assume the model inspects an original data image and its augmented variants, then all augmented instances are viewed as positive samples. The relationship vectors $\mathbf{r}_i$ (Eq.~\ref{eq:representation}), which show the similarity between all instances stored in the memory, should also be similar to the initial data points, compared to those from other instances in the same batch. In Eq.~\ref{eq:invar_prob}, the probability of an augmented instance that is correctly identified as class-$i$ is denoted as $\mathbf{\bar{p}}_i^i$; and that of $i$-th original instance that is falsely identified as class-$j$ ($j \neq i$),  $\mathbf{\bar{p}}_i^j$. The augmentation loss is then defined to minimize any misclassification over instances in all batches (Eq.~\ref{eq:aug_loss}).
\begin{align}
&\mathbf{\bar{p}}_i^i = \frac{{\exp (\mathbf{r}_i^{\top} \mathbf{\hat{r}}_i / \tau)}}{\sum_{k=1}^n \exp (\mathbf{r}_k^{\top} \mathbf{\hat{r}}_i / \tau)} \ \ \ \ \ \  \mathbf{\bar{p}}_i^j = \frac{{\exp (\mathbf{r}_j^{\top} \mathbf{r}_i / \tau)}}{\sum_{k=1}^n \exp (\mathbf{r}_k^{\top} \mathbf{r}_i / \tau)},\ \ j \neq i  \label{eq:invar_prob}
\end{align}
\vspace{-2mm}
\begin{align}
L_{aug} = -\sum_{i} \sum_{j \neq i} \log (1 - \mathbf{\bar{p}}_i^j) -\sum_{i} \log (\mathbf{\bar{p}}_i^i) \label{eq:aug_loss}
\end{align}

%Clustering is one of the frequently used strategies for unsupervised feature learning. Many of those previous methods~\cite{caron2018deep,xie2016unsupervised} use the ground truth number of clusters and adjust the network to learn mapping, which optimizes the objective of clustering. Although these deep embedding networks are useful to capture sophisticated features of input data, decision boundaries can be affected mainly by initial randomness, and overfitting to misleading characteristics can occur during the training step~\cite{zhang2017deep, ghasedi2017deep}. To solve this problem,~\cite{huang2019unsupervised} 
\section{Experiments}
We first enumerate the Super-AND model with different backbone networks on both coarse-grained and fine-grained benchmarks. Our ablation study helps speculate which components of the model are critical. We also compare the proposed model with the original AND, which our Super-AND extended from. Finally, we test Super-AND's potential to assist in semi-supervised learning tasks.

\subsection{Implementation details}

\textbf{Datasets.} A total of six datasets were utilized. Three are coarse-grained: (1) \textit{\textbf{CIFAR-10}}~\cite{krizhevsky2009learning} has 10-class images of 32 $\times$ 32 pixels. (2) \textit{\textbf{CIFAR-100}} contains images in CIFAR-10 and has 100 classes. (3) \textit{\textbf{SVHN}}~\cite{netzer2011reading} (Street View House Numbers) is a real-world dataset of 10-class images of 32$\times$32 pixels. The other three are fine-grained: (4) \textit{\textbf{Stanford Dogs (SD)}}~\cite{KhoslaYaoJayadevaprakashFeiFei_FGVC2011} contains dog images of 120 breeds, (5) \textit{\textbf{CUB-200}}~\cite{WelinderEtal2010} is from Caltech containing bird images of 200 species, and (6) \textit{\textbf{STL-10}}~\cite{coates2011analysis} contains 10-class images of 96$\times$96 pixels such as airplanes and birds. The last dataset (6) is used only for qualitative analysis. \\
%\smallskip

\noindent 
\textbf{Training.} We used AlexNet~\cite{krizhevsky2012imagenet} and ResNet18~\cite{he2016deep} as backbone networks. Hyper-parameters were tuned in the same manner as in the AND paper~\cite{huang2019unsupervised} unless explicitly mentioned. We used SGD with Nesterov momentum 0.9 for the optimizer. We fixed the learning rate as 0.03 for the first 80 epochs, and scaled-down by 0.1 every 40 epochs. Neighborhood size $k$ for AND-loss was set to 1, the batch size to 128, and the model was trained in 5 rounds and 200 epochs per round. Weights for the UE-loss $w(t)$ (Eq~\ref{eq:w}) were initialized from 0 and increased by 0.2 every 80 epochs. For the Aug-loss, we used four types: Resized Crop, Grayscale, ColorJitter, and Horizontal Flip. Horizontal Flip was not used in the case of the SVHN dataset because this dataset includes digit images. Update momentum $\eta$ of the exponential moving average for memory bank was set to 0.5. \\
%\smallskip

\noindent
\textbf{Evaluation.} Following the method from~\cite{wu2018unsupervised}, we used the weighted $k$-NN algorithm for an image classification task. Top $k$-nearest neighbors $\mathcal{N}_{top}$ were retrieved and used to predict the final outcome in a weighted fashion. We set $k=200$ and the weight function for each class $c$ as $\sum_{i \in \mathcal{N}_{top}} \exp (\mathbf{v}_i^{\top}M / \tau) \cdot \mathbf{1}(c_i = c)$ with $\tau = 0.07$, where $c_i$ is the class index for $i$-th instance. Top-1 classification accuracy was used for evaluation.

\subsection{Results \& Component analysis}

\textbf{Baseline models.} 
We adopt the following six competitive baselines for comparison: (1) \textit{Split-Brain}~\cite{zhang2017split}, (2) \textit{Counting}~\cite{noroozi2017representation}, (3) \textit{DeepCluster}~\cite{caron2018deep}, (4) \textit{Instance}~\cite{wu2018unsupervised}, (5) \textit{ISIF}~\cite{ye2019unsupervised}, and (6) \textit{AND}~\cite{huang2019unsupervised}, which also used the same backbone networks. \\
%\smallskip

\noindent\textbf{Coarse-grained evaluation.} Table~\ref{table:coarsegrained} describes the image classification performance of seven models, including the proposed Super-AND on three coarse-grained datasets: CIFAR-10, CIFAR-100, and SVHN. Super-AND outperforms competitive baselines on all datasets except for one case; ISIF performs the best on CIFAR-100 with AlexNet. For all other cases, Super-AND performs the best. We notice that the performance gain is larger when ResNet18 is used as the neural network architecture than for AlexNet. This finding indicates that Super-AND brings more substantial advantage for stronger CNN architectures such as ResNet18. \\

%One notable observation is that the difference between previous models and super-AND is mostly larger in the case of ResNet18 than for AlexNet. These results reveal that our model is superior to other methods and may indicate that our methodology can give more benefits to stronger CNN architectures. \\

\begin{table}[t!]
\caption{$k$-NN Evaluation on coarse-grained datasets. Results marked as $^*$ are borrowed from the previous work~\cite{huang2019unsupervised,ye2019unsupervised}.}
\captionsetup{margin=0.4cm}
\centering
\begin{tabular}{ccccccc}
\hline
Dataset     & \multicolumn{2}{c}{CIFAR-10} & \multicolumn{2}{c}{CIFAR-100} & \multicolumn{2}{c}{SVHN} \\ %\hline
(Network)     & ResNet18      & AlexNet      & ResNet18       & AlexNet      & ResNet18    & AlexNet    \\ \hline
Split-Brain$^*$ &        -      & 11.7         & -              & 1.3          & -           & 19.7       \\
Counting$^*$    &       -        & 41.7         & -               & 15.9         & -            & 43.4       \\
DeepCluster$^*$ &       67.6        &    62.3          &      -          &  22.7            &       -      & 84.9           \\
Instance    &         80.8      &    60.3          &       50.7         &  32.7            &    93.6         & 79.8           \\
ISIF         &      83.6         &    74.4          &      54.4         &  \textbf{44.1}   & 91.3         &   89.8        \\
AND         &       86.3        &     74.8         &        57.2        &  41.5            &        94.4     & 90.9           \\ 
 %\hline
\textbf{Super-AND}   &      \textbf{89.2}     &    \textbf{75.6}       &        \textbf{61.5}        &   42.7            &  \textbf{94.9}           &  \textbf{91.9}  \\ \hline
\end{tabular}
\label{table:coarsegrained}
\end{table}

\noindent\textbf{Fine-grained evaluation.} 
The examination on fine-grained datasets requires the ability to discriminate subtle differences across classes. Table~\ref{table:fine_grained} shows that Super-AND outperforms the competitors by a large margin on more challenging fine-grained recognition tasks. We excerpted results for Instance and AND from existing studies, and the backbone network was ResNet18. \\
%\smallskip

\noindent\textbf{Backbone network.} The choice of the backbone neural network architecture affects the overall performance. Test of AlexNet, ResNet18, and ResNet101 on the CIFAR-10 dataset, shown in Table~\ref{table:backbone}, reveals that our method performs better under stronger networks. This implies the possibility for an even higher gain under future neural network architectures. \\
%\smallskip

\noindent\textbf{Ablation study.} To examine the compactness, we removed each of the following components in turn and compared the full Super-AND model along with the following variants: (1) without the unification entropy loss, (2) without the Sobel filter, (3) without the augmentation loss. Table~\ref{table:ablation} displays the evaluation results based on the CIFAR-10 dataset and the ResNet18 backbone network. Every component is shown to contribute to the performance increase. Among the three, the largest performance gain is observed from the Sobel filter that is used for edge detection by removing color information in images.

\begin{table}[t!]
\captionsetup{margin=0.2cm}
\vspace{-2mm}
    \begin{minipage}[t]{0.38\textwidth}
    \centering
        \caption{$k$-NN evaluation on fine-grained datasets. }   
        \begin{tabular}{ccc}
            \hline
            Dataset & SD & CUB-200 \\ \hline
            Instance & 27.0 & 11.6 \\
            ISIF & 31.4 & 13.2\\
            AND        & 32.3 & 14.4 \\ %\hline
            \textbf{Super-AND}  &  \textbf{39.0} & \textbf{17.6} \\ \hline
        \end{tabular}
        \label{table:fine_grained}     
    \end{minipage}
    \begin{minipage}[t]{0.3\textwidth}
    \centering
        \caption{Backbone test on CIFAR-10.}
        \begin{tabular}{cc}
            \hline
            Network &  Accuracy\\ \hline
            AlexNet &  75.6\\
            ResNet18 & 89.2 \\
            \textbf{ResNet101} & \textbf{90.5} \\ \hline
        \end{tabular}
        \label{table:backbone} 
    \end{minipage}
    \begin{minipage}[t]{0.29\textwidth}
    \centering
        \caption{Ablation study on CIFAR-10.}
        \begin{tabular}{cc}
            \hline
            Network &  Accuracy\\ \hline
            \textbf{Full} & \textbf{89.2} \\
            Without UE &  88.7\\
            Without Aug &  88.5\\
            Without Sobel & 88.3 \\ \hline         
        \end{tabular}
        \label{table:ablation} 
    \end{minipage}
    \vspace{-3mm}
\end{table}

\subsection{Comparison to the AND model}

\begin{figure}[t!]
\centering
\captionsetup{margin=0.1cm}
\begin{minipage}{0.45\textwidth}
    \includegraphics[width=\textwidth]{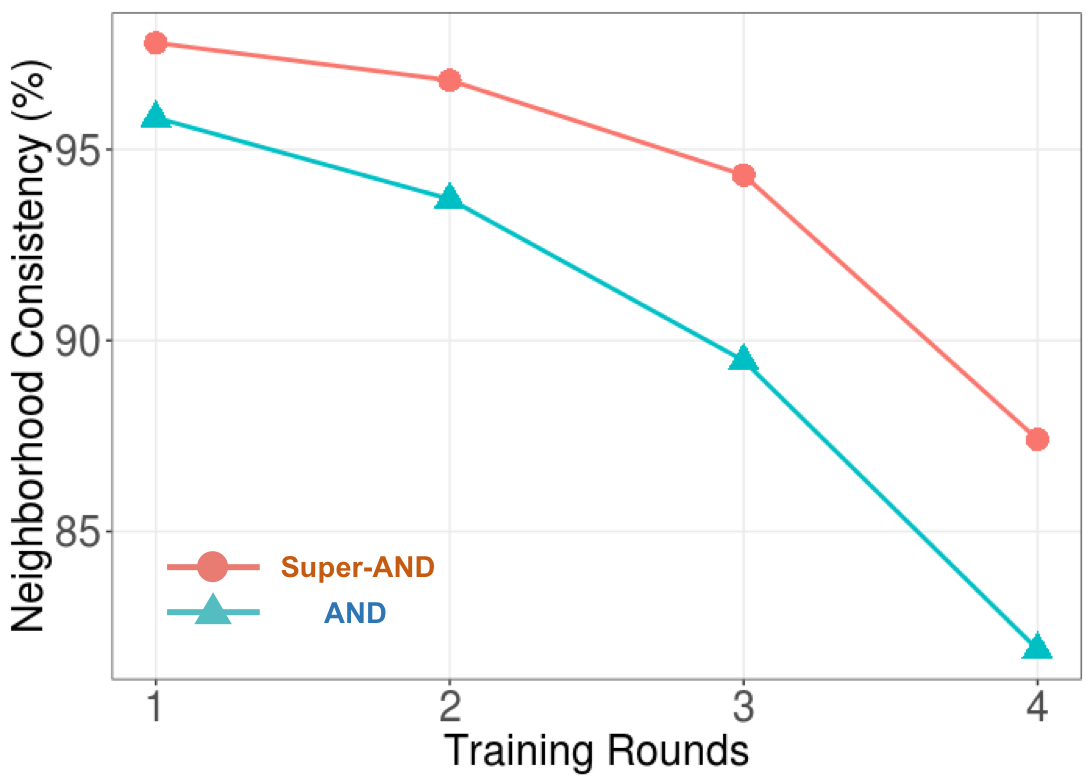}
    \caption{Neighborhood consistency over training rounds.} 
    \label{fig:consistency1} 
\end{minipage}%
\hspace{0.3cm}
\begin{minipage}{0.45\textwidth}
    \includegraphics[width=\textwidth]{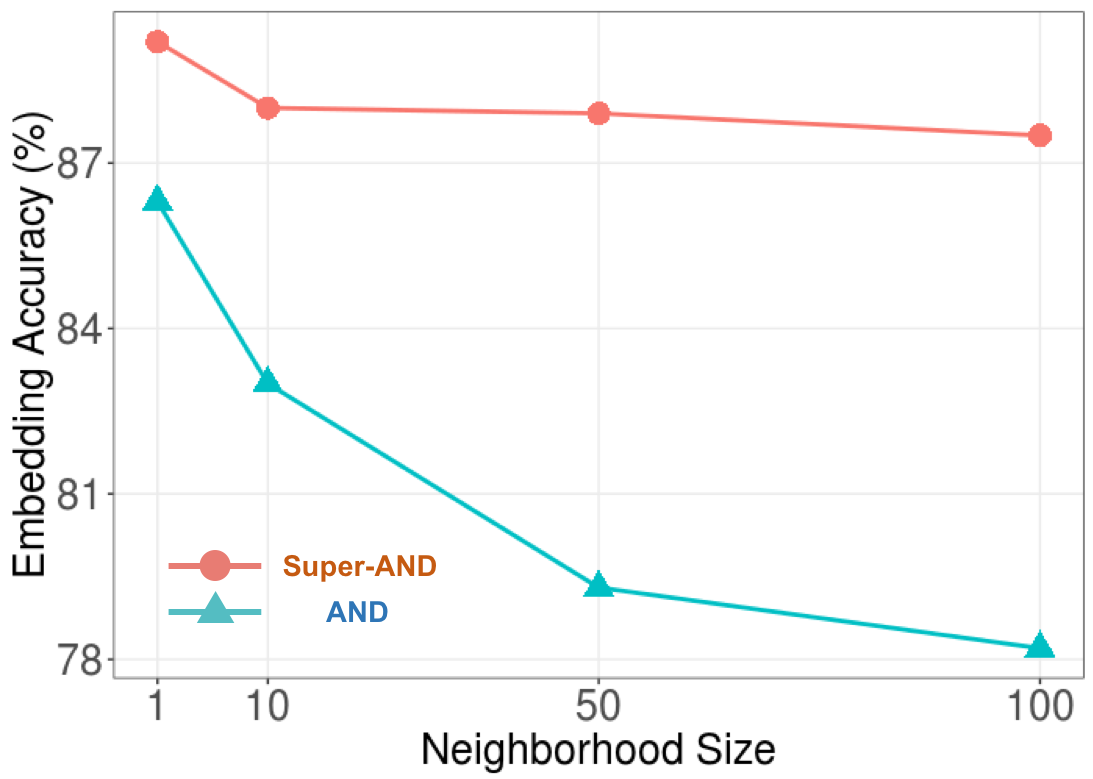}
    \caption{Effect of neighborhood size $k$ to embedding accuracy.} 
    \label{fig:consistency2}
\end{minipage}
%\vspace{2mm}
\end{figure}

\textbf{Embedding quality analysis.} 
We investigated the embedding quality by evaluating the class consistency of selected neighborhoods. Cheat labels are used to check whether neighborhood pairs come from the same class. Since both algorithms increase the selection ratio every round when gathering the part of discovered neighborhoods, the consistency of selected neighborhoods will naturally decrease. This relationship is drawn in Fig.~\ref{fig:consistency1}. The reduction, in contrast, is less significant for Super-AND; our model maintains high-performance throughout the training rounds.
%\smallskip

We evaluated the effect of neighborhood size $k$ in both algorithms (Eq.~\ref{eq:3}). Neighborhood size controls the number of data points within each cluster, and thus, the labeling consistency of each cluster is affected by this hyper-parameter. Generally, a bigger neighborhood produces poorer results due to the noise from a wider set of images as well as initial randomness. However, Fig.~\ref{fig:consistency2} shows that Super-AND model is robust to such noise even when neighborhood size reaches to a large value (i.e., $k = 100$).\\
%\smallskip

\setlength\belowcaptionskip{-3ex}

\begin{figure}[ht!]
\centering
\includegraphics[width=1\textwidth]{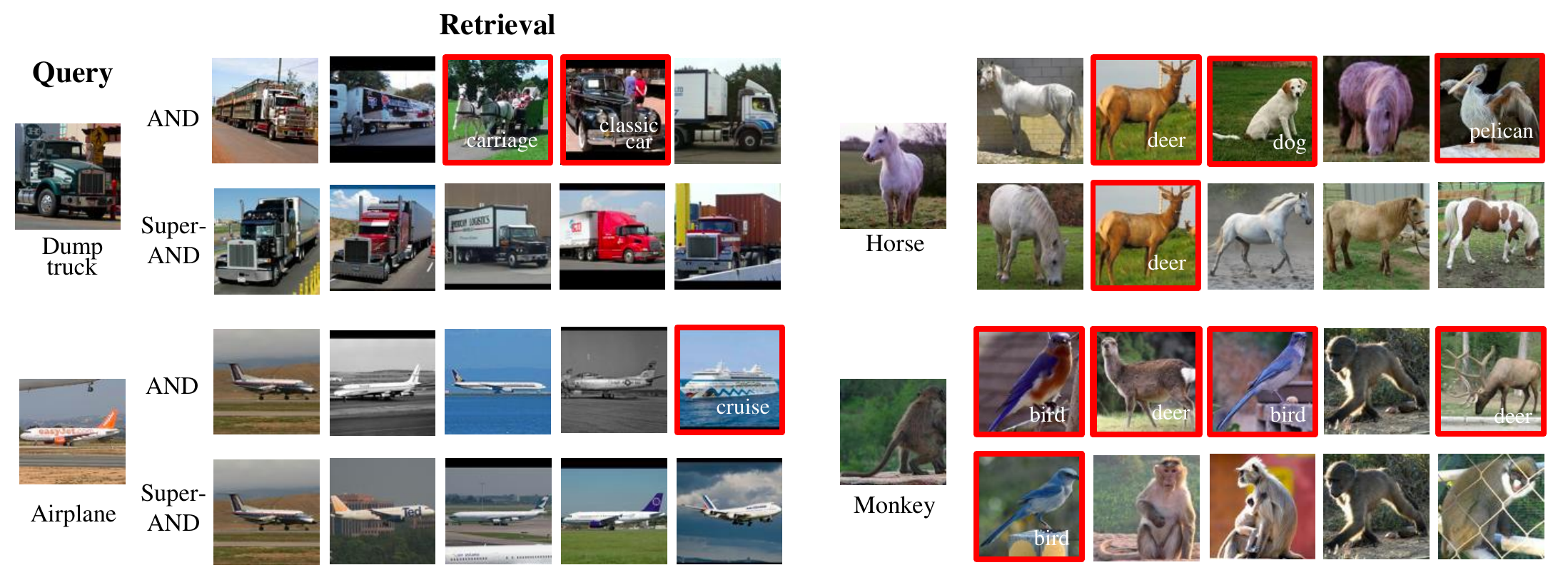}
\caption{The nearest-neighbor retrievals of example queries from STL-10. The upper retrieval row from every query shows the results from AND, and the lower ones are from Super-AND. The left-side results are successful cases for both models, and the right-side results are failure cases. Images with surrounding red frames indicate the wrongly retrieved negative samples.} 
\label{fig:qualitave1} 
\vspace{4mm}
\end{figure}

\noindent\textbf{Qualitative study.} Fig.~\ref{fig:qualitave1} illustrates the top-5 nearest retrievals of AND (i.e., upper rows) and Super-AND (i.e., lower rows) based on the STL-10 dataset. The example queries show dump trucks, airplanes, horses, and monkeys. Red framed images, which indicate negative samples (or wrong class labels), appear more frequently for AND than Super-AND. Clusters in Super-AND are robust to misleading color information and better recognizing the object shapes. For example, in the `Airplane' query, pictures retrieved from Super-AND are consistent in shape, whereas the AND result falsely includes a cruise picture. Also, in other examples, Super-AND is more flexible in allowing various colored objects such as a red dump truck or a spotted horse.

%\noindent\textbf{Qualitative study.} Fig.~\ref{fig:qualitave1} illustrates the top-5 nearest retrievals of AND (i.e., upper rows) and Super-AND (i.e., lower rows) based on the STL-10 dataset. The example queries shown are dump trucks, airplanes, horses, and monkeys. Images with red frames, which indicate negative samples, appear more frequently for AND than Super-AND. This finding implies that Super-AND excels in capturing the class information compared to AND. Its clusters are robust to misleading color information and well recognize the shape of objects within images. For example, in the case of the airplane query, pictures retrieved from Super-AND are consistent in shape, while the AND results confuse a cruise picture as an airplane. The color composition in Super-AND is also more flexible and can find a red dump truck or a spotted horse, as shown in the examples.

\subsection{Application to semi-supervised learning}

As a practical application, we demonstrate the potential for Super-AND to be used as a pre-training step for well-known semi-supervised learning tasks. We empirically confirm the usefulness by comparing the performance of several sole baseline algorithms (Supervised, $\Pi$-model~\cite{laine2016temporal}, and Mean-Teacher~\cite{tarvainen2017mean}) and that of the same algorithms with pre-trained by the proposed model. ``Supervised'' in this experiment refers to using only labeled data to train the neural network. CIFAR-10 with 4000 labels is selected for evaluation, and WideResNet28-2 is selected as a backbone network in all algorithms for a fair comparison. A breakthrough of performance is reported in Table~\ref{table:ssl} for all three cases, implying that semi-supervised learning with pre-trained Super-AND can be a practical approach for leveraging both unlabeled and labeled data when labels are scarce.

\begin{table}[!t]
\caption{Accuracy on 4000 labeled CIFAR-10. Semi-supervised learning with pre-trained network achieves improved performance. ($\Pi$ : $\Pi$-model, MT : Mean-Teacher)}
\centering
\begin{tabular*}{\textwidth}{@{\extracolsep{\fill}}ccccccc}
\hline
Dataset & Supervised & $\Pi$ & MT & Supervised (our)& $\Pi$ (our) & MT (our) \\ \hline
CIFAR-10 & 79.65     &  83.76  &   84.23 &  \textbf{84.72} & \textbf{85.23}  & \textbf{87.63}              \\ \hline
\end{tabular*}
\label{table:ssl} 
%\vspace{0mm}
\end{table}

\section{Conclusion}
This paper presented Super-AND, a comprehensive technique for unsupervised learning of deep embedding, which showed excellent potential for common computer vision tasks. Super-AND combines advantages of the existing techniques (i.e., sample specificity, clustering, anchor neighborhood discovery, data augmentation), and newly proposes a novel loss (UE-loss) that collects nearby data points nearby even in a low-density space. This method excelled in image classification tasks with both coarse-grained and fine-grained datasets, against all existing models. Implications of this research are enormous; as deep embedding learning advances, unsupervised learning approaches becomes an economically viable option in scenarios where labels are costly to generate. Our method can further assist the current semi-supervised tasks when used in the pre-training step. 
%
%This paper presents Super-AND, a holistic technique for unsupervised embedding learning. Besides the synergetic advantage combining existing methods brings, the newly proposed UE-loss that groups nearby data points even in a low-density space while maintaining invariant features via data augmentation. The experiments with both coarse-grained and fine-grained datasets demonstrate our model's outstanding performance against the state-of-the-art models. Moreover, we have shown that our Super-AND model has great potential to benefit well-known semi-supervised classification tasks. The high accuracy achieved by Super-AND makes the unsupervised learning approach an economically viable option where labels are costly to generate. 
\newpage

%Our efforts to advance unsupervised embedding learning directly benefit future applications that rely on various image clustering tasks. 

%
% ---- Bibliography ----
%
% BibTeX users should specify bibliography style 'splncs04'.
% References will then be sorted and formatted in the correct style.
%
\bibliographystyle{plain}
\bibliography{pakdd2020_conference.bib}
\end{document}